%% file: BioPose.tex
\documentclass[10pt,twocolumn,letterpaper]{article}

\usepackage{wacv}              

\usepackage{graphicx}
\usepackage{amsmath}
\usepackage{amssymb}
\usepackage{booktabs}

\usepackage{amsmath}  
\usepackage{graphicx}  
\usepackage{array}  
\usepackage{booktabs}  
\usepackage[table,xcdraw]{xcolor}  
\usepackage{makecell}  
\usepackage{caption}  
\usepackage{multirow}  
\usepackage{adjustbox}  
\usepackage{colortbl}  
\usepackage{pifont}  
\usepackage{enumitem}
\usepackage{amssymb, amsthm}
\setitemize{noitemsep,topsep=0pt,parsep=0pt,partopsep=0pt}
\newcommand{\partitle}[1]{\smallskip \noindent \textbf{#1}.}
\usepackage{float}

\usepackage[pagebackref,breaklinks,colorlinks]{hyperref}

\usepackage[capitalize]{cleveref}
\crefname{section}{Sec.}{Secs.}
\Crefname{section}{Section}{Sections}
\Crefname{table}{Table}{Tables}
\crefname{table}{Tab.}{Tabs.}
\usepackage[accsupp]{axessibility}

\def\wacvPaperID{2788} 
\def\confName{WACV}
\def\confYear{2025}

\begin{document}
\title{BioPose: Biomechanically-accurate 3D Pose Estimation from Monocular Videos}
\author{
Farnoosh Koleini\thanks{Equal contribution.}, Muhammad Usama Saleem$^{\ast}$, Pu Wang, Hongfei Xue, Ahmed Helmy, Abbey Fenwick\\
University of North Carolina at Charlotte\\
{\tt\small \{fkoleini, msaleem2, pu.wang, hongfei.xue, ahmed.helmy, afenwick\}@charlotte.edu}
}


\maketitle


\begin{abstract}

Recent advancements in 3D human pose estimation from single-camera images and videos have relied on parametric models, like SMPL. However, these models oversimplify anatomical structures, limiting their accuracy in capturing true joint locations and movements, which reduces their applicability in biomechanics, healthcare, and robotics. Biomechanically accurate pose estimation, on the other hand, typically requires costly marker-based motion capture systems and optimization techniques in specialized labs. To bridge this gap, we propose \textit{BioPose}, a novel learning-based framework for predicting biomechanically accurate 3D human pose directly from monocular videos. BioPose includes three key components: \textbf{a Multi-Query Human Mesh Recovery model (MQ-HMR)}, \textbf{a Neural Inverse Kinematics (NeurIK)} model, and \textbf{a 2D-informed pose refinement technique}. MQ-HMR leverages a multi-query deformable transformer to extract multi-scale fine-grained image features, enabling precise human mesh recovery. NeurIK treats the mesh vertices as virtual markers, applying a spatial-temporal network to regress biomechanically accurate 3D poses under anatomical constraints. To further improve 3D pose estimations, a 2D-informed refinement step optimizes the query tokens during inference by aligning the 3D structure with 2D pose observations.  Experiments on benchmark datasets demonstrate that BioPose significantly outperforms state-of-the-art methods. Project website: \url{https://m-usamasaleem.github.io/publication/BioPose/BioPose.html}.

\end{abstract}


\begin{figure}[htbp]
    \centering
    \includegraphics[width=0.45\textwidth]{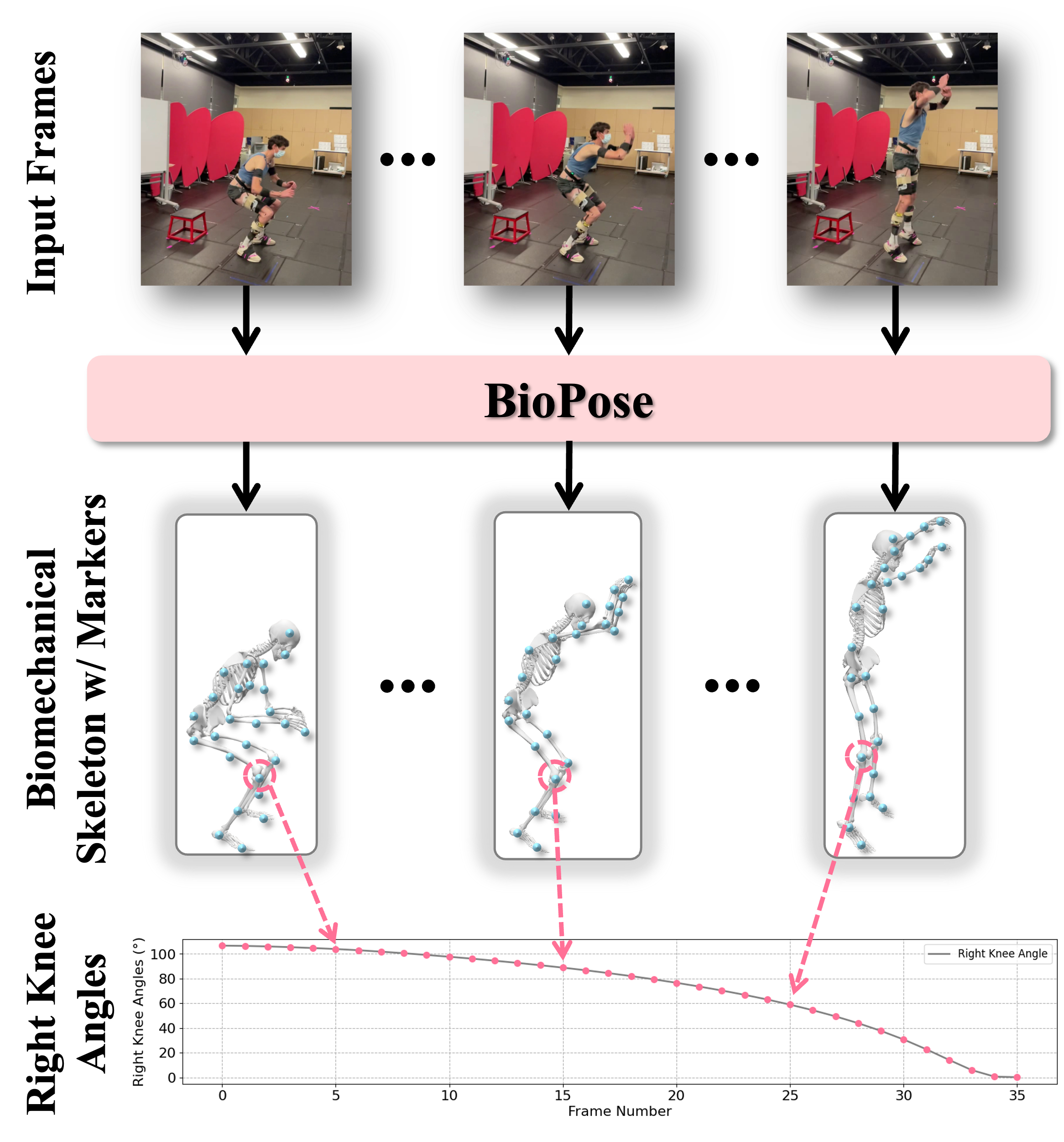}
    \caption{BioPose is a framework for biomechanically accurate 3D pose estimation from monocular videos. It first integrates Multi-Query Human Mesh Recovery (MQ-HMR), which leverages multi-scale image features for precise 3D mesh reconstruction, secondly with Neural Inverse Kinematics (NeurIK), ensuring biomechanical constraints for anatomically valid 3D poses.}
    \label{fig:biopose}
    \vspace{-12pt}
\end{figure}


\vspace{-8pt}
\section{Introduction}
\vspace{-2pt}
Biomechanically-accurate 3D human pose estimation (BA-HPE) refers to the process of predicting a human's 3D body posture, including joint positions and movements, in a way that aligns with the actual anatomical and physical constraints of the human skeletal system. BA-HPE plays a crucial role in fields like physical therapy, sports science, and ergonomics \cite{uhlrich2023opencap, ji2023using, pagnon2021pose2sim}, where precise modeling of human motion is necessary for injury prevention, rehabilitation, or performance analysis. It is also useful in robotics, animation, and human-computer interaction \cite{zhu2023human,liu2023task, zibaeirad2024comprehensive, zibaeirad2024vulnllmeval, abshari2024llm}.

Traditionally, marker-based motion capture systems have been the gold standard for obtaining biomechanically accurate 3D pose data. These systems utilize multiple cameras to track reflective markers placed on the human body in controlled laboratory environments\cite{sholukha2013model, seth2018opensim}. The captured marker data is then processed using sophisticated biomechanical optimization systems, such as OpenSim \cite{verma2020opensim},  which often requires skilled professionals to configure and refine.
Although these marker-based systems provide high accuracy, they are expensive, labor-intensive, and impractical for use outside of specialized labs, particularly in dynamic, real-world environments.  



On the other hand, significant advancements have been made in 3D pose estimation from monocular videos (single-camera footage) \cite{goel2023humans, dwivedi2024tokenhmr}, where deep neural networks are used to infer 3D poses, specifically, the rotation angles of body joints, from 2D image sequences, leveraging parametric human body models like SMPL \cite{loper2015smpl,keller2023skin}. While monocular pose estimation offers greater accessibility compared to traditional marker-based systems, it faces notable limitations due to the anatomical simplifications inherent in body models like SMPL. These body models are designed to produce visually plausible poses, but fail to achieve biomechanical accuracy, particularly in joint positioning and skeletal movements.

To address these challenges, we propose \textit{BioPose}, a novel learning-based framework for biomechanically accurate 3D human pose estimation from monocular videos. BioPose consists of three core components: a multi-query human mesh recovery model (MQ-HMR), a neural inverse kinematics (NeurIK) model, and a 2D-informed pose refinement technique. The \textit{MQ-HMR} model utilizes a multi-query deformable transformer to extract fine-grained, multi-scale image features from monocular video frames, enabling precise recovery of a 3D human mesh. This model simultaneously estimates 3D body pose, shape parameters, and camera parameters, resulting in a accurate and detailed body mesh. In the next stage, the \textit{NeurIK} model treats the recovered mesh vertices as virtual markers and uses a spatial-temporal network to regress biomechanically accurate 3D poses. This process is guided by the biomechanical skeleton model with anatomical realism, ensuring the predicted poses adhere to human biomechanics, such as anatomical locations and degrees of freedom of the joints. To further enhance accuracy, BioPose introduces a \textit{2D-informed refinement step} during inference, optimizing pose queries in latent space to align the predicted 3D poses with 2D pose cues from the video. This refinement corrects 3D-to-2D projection discrepancies, ensuring both visual coherence and biomechanical precision. Our key contributions are summarized as follows:
\begin{itemize}
    \item We introduce BioPose, a novel learning-based framework for biomechanically accurate 3D pose estimation from monocular videos, offering performance comparable to traditional marker-based optimization methods while retaining the convenience and accessibility of monocular learning-based approaches.
    \item We propose a MQ-HMR model for accurate 3D mesh reconstruction and virtual marker tracking along with the NeurIK module that incorporates a biomechanical skeleton model to ensure anatomically valid 3D poses from monocular videos.
    \item We develop a novel 2D-informed refinement technique that further enhances 3D pose estimation via inference-time optimization that aligns the predicted 3D structure with 2D pose cues.
    \item Extensive experiments demonstrate that MQ-HMR model outperforms state-of-the-art methods in monocular human mesh recovery from single images and BioPose system achieves the very competitive performance, compared with golden-standard multi-camera marker-based techniques.
\end{itemize}

\section{Related Work}
\label{sec:formatting}

\subsection{Biomechanically-accurate 3D Pose Estimation}

The gold standard for biomechanically accurate 3D pose estimation combines multi-camera, marker-based tracking systems with biomechanical optimization tools. This process involves three main steps: first, retro-reflective markers are attached to the subject’s body and tracked using synchronized infrared cameras. Next, a calibration pose (T-pose) is captured, allowing tools like OpenSim \cite{verma2020opensim} to scale a biomechanical skeleton model to the subject’s anatomy. Finally, the subject performs the target motion, and the marker data, along with the scaled skeleton model, is used to compute 3D joint rotation angles through inverse kinematics optimization.

To reduce the cost of marker-based systems, markerless multi-camera approaches like OpenCap \cite{uhlrich2023opencap, gozlan2024opencapbench} and Pose2Sim \cite{pagnon2021pose2sim} have emerged. These methods use 2D pose estimation algorithms like OpenPose \cite{cao2017realtime} to detect keypoints from video frames, triangulating them from multiple camera views to reconstruct 3D positions. However, the resulting keypoints are sparse and anatomically imprecise, requiring post-processing techniques like keypoint augmentation \cite{park2020data} to improve anatomical accuracy. These systems also require tedious camera calibration and synchronization, along with optimization tools for final 3D pose estimation.
To address these limitations, D3KE \cite{bittner2022towards} was recently proposed, using deep neural networks to directly regress biomechanically accurate 3D poses from monocular videos. However, D3KE struggles with generalization due to limited paired training data. In contrast, BioPose leverages a novel MQ-HMR model to generate robust 3D human meshes from in-the-wild videos, serving as virtual markers. These markers, combined with our biomechanics-guided NeurIK model and 2D-informed inference-time optimization, yield highly generalizable 3D pose estimations with strong anatomical accuracy.

\subsection{HMR from Monocular Images} 
Human Mesh Recovery (HMR) from monocular images has evolved significantly, focusing on estimating 3D meshes from single 2D images.  Early approaches mainly leverage convolutional neural networks to directly regress the parameters of parametric human model SMPL \cite{loper2015smpl}, from images \cite{choi2022learning, kocabas2021pare, kanazawa2018end} and videos \cite{cho2023video, choi2021beyond, kanazawa2019learning}. More recently, vision transformers \cite{alexey2020image} have been adopted for HMR tasks, yielding state-of-the-art results. Models like HMR 2.0 \cite{goel2023humans} and TokenHMR \cite{dwivedi2024tokenhmr} exploit the transformer's ability to model long-range dependencies, allowing them to better capture relationships between different body parts and improve mesh reconstruction accuracy. Besides human 3D mesh, these HMR methods also can produce the estimated 3D pose (joint rotation angles) of its underlying skeleton structure. However, such skeleton is physically inaccurate, inherently leading to inaccurate pose estimations. To address these challenges, our BioPose framework utilizes the SMPL mesh vertices as the virtual tracking markers because SMPL mesh is designed to accurately capture the deformable 3D body surface. Using these virtual markers as inputs, NeurIK model is trained to learn the 3D pose parameters of a biomechanical skeleton model with anatomical details. Since the pose estimation accuracy depends on the mesh reconstruction quality, we propose MQ-HMR model leads to improved mesh recovery results compared with SOTA solutions by harvesting multi-scale image features via multi-query deformable transformer along 2D-informed query optimization at inference.

\section{Proposed Method: BioPose} 

The goal of BioPose is to predict biomechanically accurate 3D human poses directly from monocular videos. As shown in Figure \ref{fig:biopose_overview}, BioPose has three core components. The MQ-HMR model uses a multi-query deformable transformer decoder to extract multi-scale image features from vision transformer encoder, enabling precise recovery of 3D human meshes (Section \ref{sec:MQ-HMR}). These meshes are then used by the NeurIK model, which treats the mesh vertices as virtual markers, applying a spatial-temporal network to infer biomechanically accurate 3D poses while maintaining anatomical constraints (Section \ref{sec:NeurIK}). To further improve accuracy, a 2D-informed pose refinement step aligns the 3D predictions with 2D observations, enhancing both visual coherence and biomechanical validity (Section \ref{sec:Refinement}). 


\subsection{Preliminaries}
\label{sec:Preliminaries}

\subsubsection{SMPL Human Mesh Model} We make use of the SMPL model, a differentiable parametric framework for representing human surface geometry \cite{loper2015smpl}. This model encodes the human body using pose parameters \(\theta \in \mathbb{R}^{24 \times 3}\) and shape parameters \(\beta \in \mathbb{R}^{10}\). The pose parameters \(\theta = [\theta_1, \ldots, \theta_{24}]\) include both the global orientation \(\theta_1 \in \mathbb{R}^{3}\) of the entire body and the local joint rotations \([\theta_2, \ldots, \theta_{24}] \in \mathbb{R}^{23 \times 3}\), with each \(\theta_k\) describing the axis-angle rotation of joint \(k\) relative to its parent joint in the kinematic tree. By combining these pose and shape parameters, the SMPL model produces a detailed 3D mesh \(M(\theta, \beta) \in \mathbb{R}^{3 \times N}\), where \(N = 6890\) vertices represent the surface of the body. The positions of the body joints \(J \in \mathbb{R}^{3 \times k}\) are then derived as a weighted sum of these vertices, formulated as \(J = MW\), where \(W \in \mathbb{R}^{N \times k}\) contains the predefined weights that map vertices to the corresponding joints.

 \begin{figure}[h!]
\centerline{\includegraphics[width=0.4\textwidth]{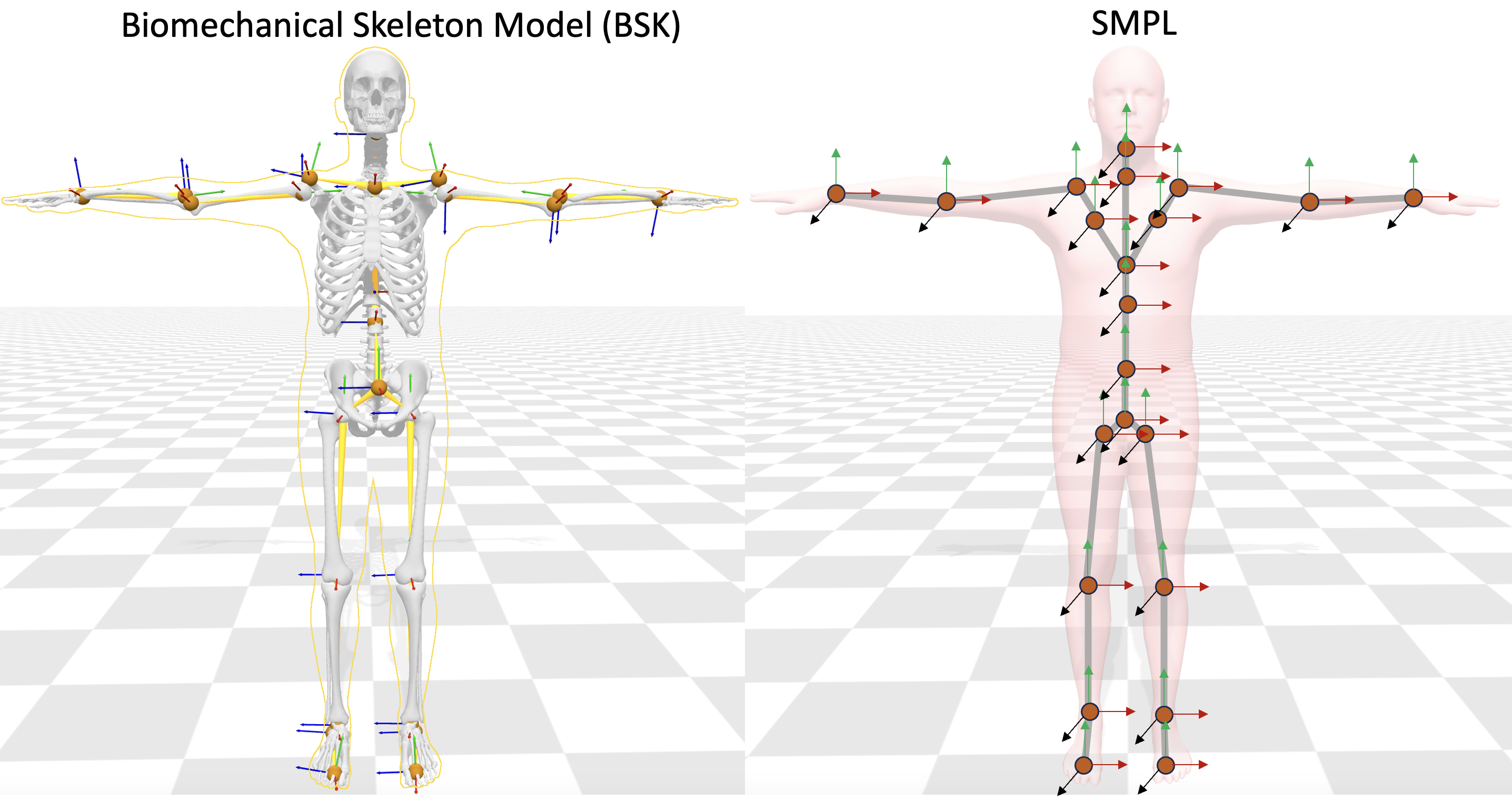}}
\caption{\textbf{Left}: Biomechanical skeleton model has anatomical details with accurate joint locations and bone orientations. \textbf{Right}: SMPL body model has deformable 3D body surface that includes an approximate skeleton geometry with inaccurate joint location and bone orientations.}
\label{fig:SMPL-BSK}
  \vspace{-20pt}
\end{figure}

\subsubsection{Biomechanical Skeleton (BSK) Model}
The BSK model, e.g., widely-adopted OpenSim models, is represented by a series of bone segments that are interconnected through movable anatomical joints, which possess anatomical movement constraints, such as Degrees of Freedom, to limit the range of motion of the respective body parts. In particular, the BSK model  generally consists of $24$ rigid bone segments, which are represented by three sets of parameters $(q^{o}, q^{r}, s)$. The anatomical joint orientation  \(q^{o} \in [q_1^{o}, \ldots, q_{24}^{o}] \) with \(q_i^{o} \in \mathbb{R}^{3}\) defines the relative orientation of each joint with respective to its parent joint along the kinematic skeleton tree. Therefore, $q^{o}$ are  determined by the anatomical structure of human skeleton.  \(q^{r} \in  [q_1^{r}, \ldots, q_{24}^{r}]\) represents the motion-induced joint rotation with \(q_i^{r} \in \mathbb{R}^{D_i}\) and $D_i \leq 3$ represents the Euler's angle rotation of joint \(i\) relative to its parent in the kinematic tree under the constraints imposed by the degree of freedom $D_i$ of each joint $i$. The bone scale \(s  \in [s_1, \ldots, s_{24}] \) with \(s_i \in \mathbb{R}^{3}\) aims to tailor the generic anatomical skeleton model (in the rest pose) by scaling each bone length and shape along with the $(x, y, z)$ axis. The scaled skeleton yields the body joints  \(p_{J} \in \mathbb{R}^{3 \times 24}\) at anatomical positions. The differences between SMPL and BSK models are shown in Fig. \ref{fig:SMPL-BSK}

\begin{figure*}[htbp]
    \centering
    \includegraphics[width=0.9\linewidth]{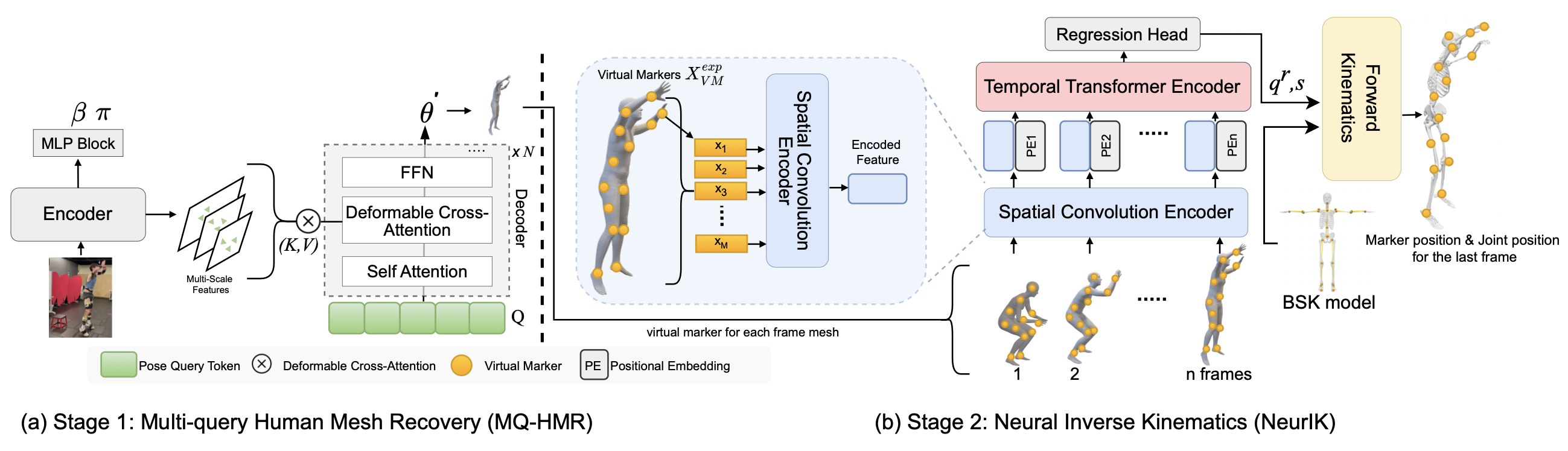}
\caption{Overview of BioPose, comprising two key components: (1) the \textbf{MQ-HMR} model, which leverages a multi-query deformable transformer decoder to extract multi-scale image features from a vision transformer, enabling precise 3D mesh recovery, and (2) the \textbf{NeurIK} model, which uses the mesh vertices as virtual markers and applies a spatio-temporal network to infer biomechanically accurate 3D poses while maintaining anatomical constraints.}

    \label{fig:biopose_overview}
\end{figure*}
\subsubsection{Optimization-based Biomechanical Kinematics}
With assistance of the BSK model, the kinematic analysis aims to find the optimal pose, i.e., joint rotation angles 
$q^{r}$, which can best fit the BSK model to the motion capture sequences. Towards this goal, a set of model markers is first attached to the bone segments in such a way that each bone segment is associated with at least $D_i$ markers to ensure the unique solutions of the derived rotation angles at joint $i$ with $D_i$ degrees of freedom. Then, a set of corresponding experiential markers is placed on the human subject. Then, the pose $q^{r}$ and bone scale $s$ can be obtained by solving an optimization problem that minimizes the distance between each experimental marker and its corresponding model marker. i.e., 
\begin{align*}
 q^{r*}, s^* =\arg\min_{ q^{r}, s}\sum_{i = 1}^{M} \left\| {f_{FK}(q^{r}, s, q^{o}, p_i)} - x_i^{exp}\right\|      
\end{align*}
where $M$ is the number of markers. $p_i \in \mathbb{R}^{3}$ denotes the position of $i$-th model marker in the local coordination system of the body segment to which it is attached.  $f_{FK}(q^{r}, s, q^{o}, p_i)$ is the forward kinematics transformation that converts the model marker $i$ from its local coordination frame to the world coordination system under the scaled skeleton with the pose of $q^{r}$. $x_i^{exp} \in \mathbb{R}^{3}$ is the position of the experiential marker $i$ in the world coordination system. We leverage  OpenSim, the classic biomechanical optimizer, and BML-MoVi dataset \cite{ghorbani2021movi} to obtain $q^{r*}, s^*$, which serve as the ground-truth data to train the NeurIK model.

\subsection{MQ-HMR} 
\label{sec:MQ-HMR}
The goal of MQ-HMR is predicting accurate virtual experiential markers, \(X_{\text{VM}}^{exp}\), from the 3D mesh, which serves as the inputs for NeurIK model. Towards this goal, MQ-HMR model consists of two key components: the image encoder and the multi-query deformable transformer decoder.

\partitle{Image Encoder} Our image encoder is based on the Vision Transformer (ViT), specifically the ViT-H/16 variant \cite{alexey2020image, dwivedi2024tokenhmr}. The encoder begins by dividing the input image into 16x16 pixel patches, which are processed through multiple transformer layers to produce a set of feature tokens that encode the visual information. To enhance this process, we implement a multi-scale feature extraction approach \cite{alexey2020image}. After generating the initial feature map, the encoder upsamples it to produce feature maps at various resolutions. These higher-resolution maps capture fine-grained visual details, such as joint positions and orientations, while lower-resolution maps preserve broader, high-level semantic information, such as the overall human skeleton structure. This multi-scale strategy allows the encoder to capture both intricate local details and global contextual information simultaneously, which is crucial for accurate pose estimation.

\partitle{Multi-Query Deformable Transformer Decoder} Building upon multi-scale feature maps, the multi-query deformable transformer decoder introduces a novel mechanism designed to recover precise 3D human poses by extracting fine-grained semantic information from diverse resolutions. The multi-query approach initializes multiple pose queries as zero pose tokens, which interact with the encoder’s multi-scale feature maps. These queries generate pose anchors crucial for accurately estimating complex human poses, especially in challenging scenarios like occlusions or ambiguous body positions. To process high-resolution feature maps efficiently, MQ-HMR incorporates a multi-scale deformable cross-attention mechanism \cite{Zhu2020DeformableDD,lin2023one}, focusing each query on a small set of sampling points near the pose anchors and dynamically adjusting attention to the most relevant regions. This optimizes both computational efficiency and accuracy, allowing the model to concentrate on critical spatial features with minimal overhead. The deformable attention mechanism (DAM) for multi-scale features is formulated as:

\begin{equation*}
\text{DAM}(\mathbf{Q}, \hat{\mathbf{r}}_k, \{F^s\}_{s=1}^S) = \sum_{s=1}^S \sum_{m=1}^M \alpha_{ksm} \cdot \mathbf{W} F^s \left( \hat{\mathbf{r}}_k + \Delta r_{ksm} \right)
\end{equation*}
Where \( \mathbf{Q} \) represents the pose token queries, \( \hat{\mathbf{r}}_k \) denotes the learnable reference points, \( \Delta r_{ksm} \) refers to the learnable sampling offsets around the reference points, \( \{ F^s \}_{s=1}^S \) are the multi-scale image features, \( \alpha_{ksm} \) are the attention weights, and \( \mathbf{W} \) is a learnable weight matrix. This deformable cross-attention strategy allows the model to capture fine-grained details while balancing computational efficiency. From this accurately reconstructed 3D mesh, virtual markers \(X_{\text{VM}}^{exp}\) are extracted, serving as input to the NeurIK module for further refinement and achieving precise biomechanical accuracy.


\partitle{MQ-HMR Losses} In line with established practices in HMR research \cite{dwivedi2024tokenhmr, goel2023humans}, we train our MQ-HMR model using a combination of losses based on SMPL parameters, 3D keypoints, and 2D keypoints. The final overall loss is:

\begin{equation}
L_{\text{total}} = L_{\text{smpl}} + L_{\text{3D}} + L_{\text{2D}}
\label{eq:total_loss}
\end{equation}
where $L_{\text{smpl}}$ minimizes the error between the predicted and ground-truth SMPL pose ($\theta$) and shape ($\beta$) parameters, $L_{\text{3D}}$ supervises the accuracy of the 3D keypoint predictions, and $L_{\text{2D}}$ enforces consistency between the projected 3D keypoints and their corresponding 2D annotations.


\subsection{NeurIK} 
\label{sec:NeurIK}

After extracting virtual markers \( \mathbf{X}_{\text{VM}}^{\text{exp}} \) from MQ-HMR, the NeurIK module processes these markers to predict biomechanically accurate 3D poses $q_r$ and bone scale $s$. To achieve this, it employs three key components: i) a Spatial Convolution Encoder to model spatial relationships among body parts, ii) a Temporal Transformer Encoder to capture dynamic motion patterns over time, and iii) multiple loss functions that incorporate biomechanical constraints from a musculoskeletal model.

\partitle{Spatial Convolution Encoder} The Spatial Convolution Encoder is designed to extract high-dimensional spatial features from a single frame. Given the 3D human mesh generated by the pre-trained MQ-HMR model, we extract \( M \) virtual markers from the mesh, where each marker \( m \) has 3D coordinates \( (x_m, y_m, z_m) \). These marker positions \( \mathbf{X}_{\text{VM}}^{\text{exp}} \in \mathbb{R}^{M \times 3} \) are first projected into a higher-dimensional space using a trainable linear projection, resulting in the spatial embedding \( \mathbf{Z}_{n_i} \in \mathbb{R}^{M \times c} \), where \( c \) is the spatial embedding dimension. To capture spatial relationships across the body, this spatial embedding is processed through a series of 1-D convolutional layers, which capture both local relationships between neighboring markers and global dependencies across the entire body structure. The output of this spatial convolution process for frame \( n_i \), \( \mathbf{Z}_{n_i} \in \mathbb{R}^{M \times c} \), represents a refined spatial feature embedding, which is passed to the Temporal Transformer Encoder for temporal modeling.

\partitle{Temporal Transformer Encoder} After encoding high-dimensional spatial features for each individual frame, the Temporal Transformer Encoder models the dependencies across the sequence of frames. For frame \( n_i \), the spatially encoded feature matrix \( \mathbf{Z}_{n_i} \in \mathbb{R}^{M \times c} \) is flattened into a vector \( \mathbf{z}_{n_i} \in \mathbb{R}^{1 \times (M \cdot c)} \). We concatenate these vectors across all \( n \) frames to form the sequence matrix \( \mathbf{Z}_{\text{seq}} \in \mathbb{R}^{n \times (M \cdot c)} \), which represents the spatial features for the entire motion sequence. To capture temporal relationships, we add a learnable temporal positional embedding \( \mathbf{PE}_n \in \mathbb{R}^{n \times (M \cdot c)} \) to the sequence matrix. The temporal transformer then applies multi-head self-attention blocks and feed-forward layers to model both short-term and long-term dependencies across frames. This allows the model to understand the progression of motion and how body parts evolve over time. The final output of the temporal transformer, \( \mathbf{Y}_n \in \mathbb{R}^{n \times (M \cdot c)} \), is used to predict key biomechanical parameters such as body scales \( \mathbf{s} \) and joint angles \( \mathbf{q}^r \). These predictions are further refined through a Forward Kinematics (FK) module to ensure biomechanically accurate marker and joint positions.


\partitle{NeurIK Losses} Spatial and temporal models are trained using multiple supervisions, including joint positions, marker positions, body scales, and joint angles. The joint positions correspond to anatomical landmarks in the musculoskeletal model, ensuring biomechanical accuracy. The overall loss function \( L \) is a weighted sum of four terms: \( L_j \) for joint positions, \( L_m \) for marker positions, \( L_s \) for body scales, and \( L_q \) for joint angles. These terms are weighted by coefficients \( \lambda_j \), \( \lambda_m \), \( \lambda_s \), and \( \lambda_q \), respectively, to control their contributions to the total loss. The \textbf{overall loss function} is defined as:
\begin{equation}
L_{neurIK} = \lambda_j L_j + \lambda_m L_m + \lambda_s L_s + \lambda_q L_q
\end{equation}
In particular, both $L_j$ and $L_m$ incorporate  biomechanical constraints during training through the forward kinematics (FK) layer. As shown in Fig. \ref{fig:biopose_overview}, the FK layer transforms the rest-pose BSK model markers and anatomical joints to the new positions according to the estimated rotation angles $q^r$ in such a way that the model markers best match the virtual experimental markers. Since we employed a full-body skeletal model from OpenSim \cite{seth2018opensim}, the FK transformation is inherently contrained by the realsitic degrees of freedom and range of motions of body joints. The details of loss functions are shown in the supplementary material.

\begin{figure}[htbp]
    \centering
    \includegraphics[width=1.0\linewidth]{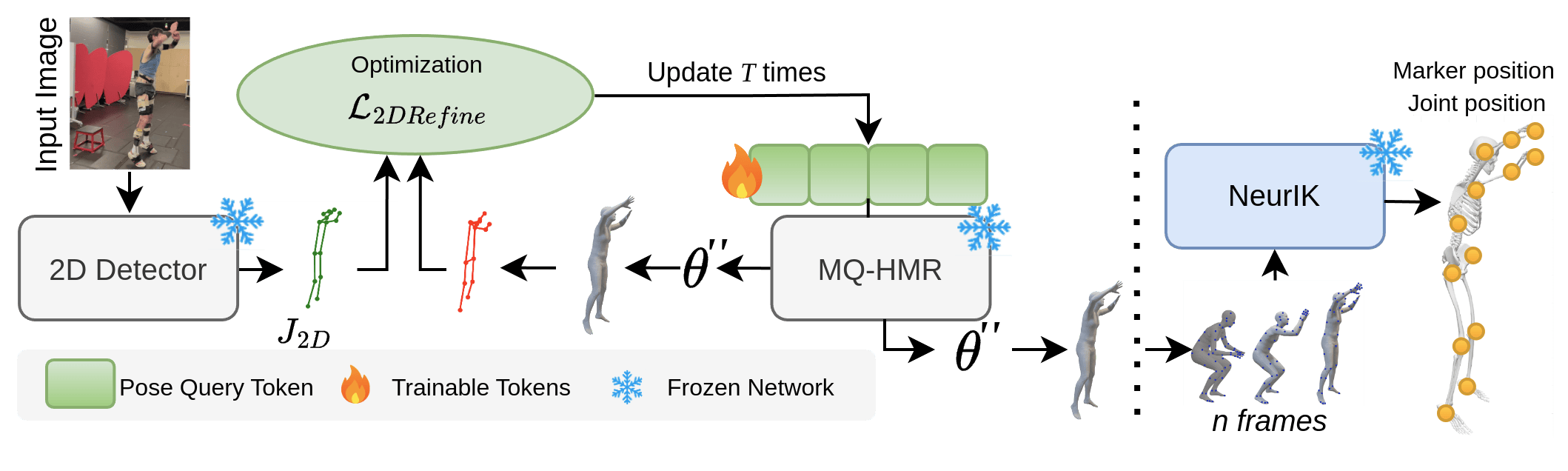}
\caption{Our \textbf{2D Pose-Guided Refinement} process fine-tunes the pose query tokens to minimize 3D reconstruction errors by aligning the 3D body mesh with 2D pose estimates, improving visual coherence and providing more reliable virtual markers for the NeurIK module to further refine 3D biomechanical pose accuracy.}
    \label{fig:biopose_testtime}
\end{figure}

\subsection{2D Pose-Informed Refinement at Inference}  
\label{sec:Refinement}
During inference time, to further reduce uncertainties in the 3D reconstruction process, we fine-tune the pose query tokens \( Q \) within the MQ-HMR model while keeping the rest of the network frozen as shown in Figure \ref{fig:biopose_testtime}. This refinement ensures alignment between the predicted 3D poses and 2D pose data, leveraging robust 2D pose detectors such as OpenPose \cite{cao2017realtime}. The process begins with the pose query tokens generated by MQ-HMR, and the initial pose parameters \( \theta' \) are derived from the output of MQ-HMR for a given input image. These tokens and parameters are iteratively adjusted to minimize discrepancies between the inferred 3D and observed 2D poses by optimizing a guidance function \( F(Q_t, J_{\text{2D}}, \theta') \). This function penalizes misalignments between the two pose domains while enforcing regularization through the following expression:

\begin{equation}
Q^+ = \arg\min_{Q_t} \left( \mathcal{L}_{2DRefine}(J_{\text{3D}}') + \lambda_{\theta'} \mathcal{L}_{\theta'}(\theta') \right)
\end{equation}
where the term \( \mathcal{L}_{2DRefine}(J_{\text{3D}}') \) aims to align the reprojected 3D joints \( J_{\text{3D}}' \) with the detected 2D keypoints \( J_{\text{2D}} \), using the following relation:

\begin{equation}
\mathcal{L}_{2DRefine}(J_{\text{3D}}') = |\Pi(K(J_{\text{3D}}')) - J_{\text{2D}}|^2
\end{equation}
where \( \Pi(K(\cdot)) \) represents the perspective projection function governed by the camera intrinsics \( K \). Simultaneously, the regularization term \( \mathcal{L}_{\theta'}(\theta') \) constrains the pose parameters \( \theta' \), ensuring that they do not diverge excessively from their initial values, thereby avoiding anatomically implausible body configurations. The pose tokens \( Q_t \) are iteratively updated at each step \( t \) via gradient-based optimization as follows:

\begin{equation}
Q_{t+1} = Q_t - \eta \nabla_{Q_t} F(Q_t, J_{\text{2D}}, \theta')
\end{equation}
where \( \eta \) determines the step size for the update, and \( \nabla_{Q_t} F(Q_t, J_{\text{2D}}, \theta') \) denotes the gradient of the objective function with respect to the pose tokens at the current iteration \( t \). This iterative refinement process, carried out over \( T \) total iterations, gradually fine-tunes the pose representation, reducing the gap between the 3D estimates and their 2D counterparts while keeping realistic pose configurations.





\section{Experiments}
\begin{table*}[h!]
\centering
\caption{MQ-HMR Reconstruction Evaluation in 3D: Reconstruction errors (measured in mm) on the Human3.6M, 3DPW, and EMDB datasets. Smaller values ($\downarrow$) represent superior performance. \underline{Underlined} values highlight the second-best performance in each column. \textcolor{blue}{Blue} highlights the improvements of our MQ-HMR approach over the second-best method. - indicates missing results.}
\vspace{-7pt}
\scalebox{0.7}{
\begin{tabular}{ll|cc|ccc|ccc}
\toprule
\multicolumn{2}{c|}{} & \multicolumn{2}{c|}{Human3.6M} & \multicolumn{3}{c|}{3DPW} & \multicolumn{3}{c}{EMDB} \\
Methods & Venue & \makecell{PA-MPJPE \\ (↓)} & \makecell{MPJPE \\ (↓)} & \makecell{PA-MPJPE \\ (↓)} & \makecell{MPJPE \\ (↓)} & \makecell{MVE \\ (↓)} & \makecell{PA-MPJPE \\ (↓)} & \makecell{MPJPE \\ (↓)} & \makecell{MVE \\ (↓)} \\
\midrule
FastMETRO \cite{cho2022cross} & \textcolor{gray}{\textit{ECCV 2022}} & 33.7 & 52.2 & 65.7 & 109.0 & 121.6 & 72.7 & 108.1 & 119.2 \\
PARE \cite{kocabas2021pare} & \textcolor{gray}{\textit{ICCV 2021}} & 50.6 & 76.8 & 50.9 & 82.0 & 97.9 & 72.2 & 113.9 & 133.2 \\
Virtual Marker \cite{ma20233d} & \textcolor{gray}{\textit{CVPR 2023}} & - & - & 48.9 & 80.5 & 93.8 & - & - & - \\
CLIFF \cite{li2022cliff} & \textcolor{gray}{\textit{ECCV 2022}} & \underline{32.7} & 47.1 & 46.4 & 73.9 & 87.6 & 68.8 & 103.1 & 122.9 \\
HMR2.0 \cite{goel2023humans} & \textcolor{gray}{\textit{ICCV 2023}} & 33.6 & \underline{44.8} & \underline{44.5} & \underline{70.0} & \underline{84.1} & \underline{61.5} & \underline{97.8} & 120.1 \\
VQ HPS \cite{fiche2023vq} & \textcolor{gray}{\textit{ECCV 2024}} & - & - & 45.2 & 71.1 & 84.8 & 65.2 & 99.9 & \underline{112.9} \\
TokenHMR \cite{dwivedi2024tokenhmr} & \textcolor{gray}{\textit{CVPR 2024}} & 36.3 & 48.4 & 47.5 & 75.8 & 86.5 & 66.1 & 98.1 & 116.2 \\
\rowcolor{gray!15} \textbf{MQ-HMR} & \textbf{\textit{Ours}} & \textbf{28.5} \textbf{\textcolor{blue}{\scriptsize{(4.2$\downarrow$)}}} & \textbf{42.5} \textbf{\textcolor{blue}{\scriptsize{(2.3$\downarrow$)}}} & \textbf{39.5} \textbf{\textcolor{blue}{\scriptsize{(5.0$\downarrow$)}}} & \textbf{69.0} \textbf{\textcolor{blue}{\scriptsize{(1.0$\downarrow$)}}} & \textbf{79.8} \textbf{\textcolor{blue}{\scriptsize{(4.3$\downarrow$)}}} & \textbf{52.1} \textbf{\textcolor{blue}{\scriptsize{(9.4$\downarrow$)}}} & \textbf{92.5} \textbf{\textcolor{blue}{\scriptsize{(5.3$\downarrow$)}}} & \textbf{98.9} \textbf{\textcolor{blue}{\scriptsize{(14.0$\downarrow$)}}} \\
\bottomrule
\end{tabular}
}
\label{tab:MQHMR_results}
\begin{flushleft}
\end{flushleft}
\vspace{-7pt}
\end{table*}

\subsection{Datasets} To train MQ-HMR, we employs a diverse collection of datasets. In line with prior research \cite{goel2023humans} and to ensure consistency in comparisons with other baselines, a diverse set of standard datasets (SD) is used for training, including Human3.6M (H36M) \cite{ionescu2013human3}, COCO \cite{lin2014microsoft}, MPI-INF-3DHP \cite{mehta2017monocular}, and MPII \cite{andriluka20142d}. For NeurIK, we utilizes BML-MoVi \cite{ghorbani2021movi} for training, which provides rich biomechanical motion capture and video data from multiple actors performing everyday activities.

\subsection{Evaluation Metrics}

The performance of MQ-HMR and NeurIK is evaluated using a set of key metrics designed for evaluating 3D human pose accuracy and biomechanical estimation.  For MQ-HMR, the metrics used include Mean Per Joint Position Error (MPJPE) and Mean Vertex Error (MVE) to evaluate the accuracy of 3D pose estimation. Additionally, we report Procrustes-Aligned MPJPE (PA-MPJPE), which measures the alignment between predicted and ground-truth poses after a rigid transformation, providing a more precise comparison of pose structure. MQ-HMR is evaluated on the Human3.6M testing split, consistent with prior studies \cite{kolotouros2019learning}, and for \emph{generalization} testing, we assess the model on challenging real-world datasets like 3DPW \cite{von2018recovering} and EMDB \cite{kaufmann2023emdb}. Importantly, \textbf{\emph{no}} training is performed on these datasets, ensuring a fair evaluation on unseen data.

For NeurIK, we employ metrics that focus on biomechanical accuracy. These include Mean Per Bony Landmarks Position Error (MPBLPE), which measures the accuracy of predicted bony landmarks against ground truth positions, and Mean Absolute Error for body scale (MAE$_{\textit{body}}$), which evaluates the correctness of predicted body segment dimensions by comparing their longest axes in millimeters. Additionally, we report Mean Absolute Error for joint angles (MAE$_{\textit{angle}}$), which assesses the precision of joint angle predictions in degrees, critical for biomechanical simulations such as joint force and muscle force analysis. To assess computational efficiency, we use the metric Average Inference Time per Image (AITI) in seconds, similar to the average inference or optimization time per image described by OpenPose \cite{cao2017realtime}. Unlike that, AITI tracks the processing time per iteration, allowing for a more detailed evaluation of our 2D Pose-Informed Refinement technique. AITI is obtained on a single  NVIDIA Quadro RTX A4000. Lower values in these metrics indicate better model performance. For NeurIK, we test the model on the BMLmovi \cite{ghorbani2021movi}, OpenCap \cite{uhlrich2023opencap}, and BEDLAM \cite{black2023bedlam} datasets. NeurIK is  \textbf{\emph{not}} trained on OpenCap and BEDLAM datasets to show its cross-dataset generalization performance.

\begin{table*}[h!]
\centering
\caption{NeurIK Biomechanical Analysis in 3D: Evaluation of biomechanical accuracy on the BML-MoVi, BEDLAM, and OpenCap datasets. Metrics include body scale error (MAE$_{\textit{body}}$), marker positions (MPBLPE), and joint angle error (MAE$_{\textit{angle}}$). Smaller values ($\downarrow$) indicate better performance. \underline{Underlined} values highlight the fourth-best performance in each column, achieved by existing HMR2.0 \cite{goel2023humans} + NeurIK (our NeurIK module combined with HMR2.0). \textcolor{blue}{Blue} highlights the improvements of our MQ-HMR + NeurIK$^\dagger$ over the fourth-best method. - indicates missing results.}

\vspace{-7pt}
\scalebox{0.7}{
\begin{tabular}{l|ccc|ccc|ccc}
\toprule
 & \multicolumn{3}{c|}{BML-MoVi Dataset} & \multicolumn{3}{c|}{BEDLAM Dataset} & \multicolumn{3}{c}{OpenCap Dataset} \\
Methods & \makecell{MAE$_{\textit{body}}$ \\ (↓)} & \makecell{MPBLPE\\ (↓)} & \makecell{MAE$_{\textit{angle}}$\\ (↓)} & \makecell{MAE$_{\textit{body}}$ \\ (↓)} & \makecell{MPBLPE\\ (↓)} & \makecell{MAE$_{\textit{angle}}$\\ (↓)} & \makecell{MAE$_{\textit{body}}$\\ (↓)} & \makecell{MPBLPE\\ (↓)} & \makecell{MAE$_{\textit{angle}}$\\ (↓)} \\
\midrule
D3KE \cite{bittner2022towards} & 5.90 & 36.98 & 3.54 & 8.26 & 39.45 & 6.72 & 7.15 & 38.62 & 5.92 \\
OpenCap Multi-Camera \cite{uhlrich2023opencap} & - & - & - & - & - & - & - & - & 4.50 \\
\midrule
HMR2.0 \cite{goel2023humans} + NeurIK (Ours) & \underline{5.86} & \underline{35.90} & \underline{3.31} & \underline{6.01} & \underline{36.62} & \underline{3.85} & \underline{5.89} & \underline{36.07} & \underline{3.41} \\
\rowcolor{gray!15} \textbf{HMR2.0} \cite{goel2023humans} + \textbf{NeurIK} (Ours) & \textbf{4.11} & \textbf{26.31} & \textbf{2.73} & \textbf{4.84} & \textbf{27.02} & \textbf{3.55} & \textbf{5.29} & \textbf{27.52} & \textbf{3.21}\\
\midrule
MQ-HMR + NeurIK (Ours)  & 5.26 & 28.39 & 3.87 & 5.62 & 30.12 & 4.16 & 5.42 & 32.17 & 4.02\\
\rowcolor{gray!15} \textbf{MQ-HMR + NeurIK}$^\dagger$ (Ours) & \textbf{3.97} \textbf{\textcolor{blue}{\scriptsize{(32\%$\downarrow$)}}} & \textbf{25.76} \textbf{\textcolor{blue}{\scriptsize{(28\%$\downarrow$)}}} & \textbf{2.84} \textbf{\textcolor{blue}{\scriptsize{(14\%$\downarrow$)}}} & \textbf{4.28} \textbf{\textcolor{blue}{\scriptsize{(29\%$\downarrow$)}}} & \textbf{26.54} \textbf{\textcolor{blue}{\scriptsize{(27\%$\downarrow$)}}} & \textbf{3.14} \textbf{\textcolor{blue}{\scriptsize{(18\%$\downarrow$)}}} & \textbf{4.87} \textbf{\textcolor{blue}{\scriptsize{(17\%$\downarrow$)}}} & \textbf{26.34} \textbf{\textcolor{blue}{\scriptsize{(27\%$\downarrow$)}}} & \textbf{3.19} \textbf{\textcolor{blue}{\scriptsize{(6\%$\downarrow$)}}} \\
\bottomrule
\end{tabular}
}
\label{tab:sota_results}
\begin{flushleft}
\footnotesize{$^\dagger$ indicates MQ-HMR with 2D Pose-Informed Refinement during inference. OpenCap method is the multi-camera markerless motion capture system. D3KE is the single-camera solution. HMR2.0 + NeurIK indicates the virtual markers come from the mesh vertices of HMR 2.0, the SOTA solution for monocular HMR.}
\end{flushleft}
\end{table*}

\subsection{Comparison to State-of-the-art Approaches}

\subsubsection{Quantitative Evaluation of MQ-HMR}

As shown  in Table \ref{tab:MQHMR_results}, MQ-HMR demonstrates significant improvements over existing methods across multiple benchmark datasets, particularly on Human3.6M, 3DPW, and EMDB. The model's performance is especially noteworthy on EMDB, which simulates complex, real-world scenarios, achieving substantial reductions in PA-MPJPE (by 9.4 mm) and MVE (by 14.0 mm), showcasing its ability to excel in intricate environments with enhanced precision. Similarly, on 3DPW, known for its in-the-wild settings, MQ-HMR further demonstrates its effectiveness by lowering MPJPE by 5.0 mm and MVE by 4.3 mm, confirming its robust performance in diverse 3D pose reconstruction tasks. The key innovation driving these improvements is MQ-HMR's use of multi-scale deformable attention, enabling simultaneous capture of fine-grained details and broader contextual information. This balanced approach enhances both precision and computational efficiency. Consequently, these results highlight MQ-HMR's ability to generalize effectively across both controlled and real-world conditions. 

\begin{table*}[htbp]
\centering
\caption{Impact of Number of Iterations on 2D Pose-Informed Refinement}
\vspace{-7pt}
\scalebox{0.7}{
\begin{tabular}{l|ccc|ccc|ccc|c}
\toprule
 & \multicolumn{3}{c|}{BML-MoVi} & \multicolumn{3}{c|}{BEDLAM} & \multicolumn{3}{c|}{OpenCap} & AITI \\
\# of Iteration & \makecell{MAE$_{\textit{body}}$} & \makecell{MPBLPE} & \makecell{MAE$_{\textit{angle}}$} & \makecell{MAE$_{\textit{body}}$} & \makecell{MPBLPE} & \makecell{MAE$_{\textit{angle}}$} & \makecell{MAE$_{\textit{body}}$} & \makecell{MPBLPE} & \makecell{MAE$_{\textit{angle}}$} & (s) \\
\midrule
1  & 4.89 & 26.23 & 3.47 & 5.61 & 27.61 & 4.11 & 5.41 & 27.82 & 3.98 & 0.293\\
5  & 4.12 & 25.92 & 2.97 & 5.13 & 27.19 & 3.45 & 5.33 & 27.15 & 3.97 & 0.405 \\
10 & \textbf{3.97} & \textbf{25.76} & \textbf{2.84} & \textbf{4.28} & \textbf{26.54} & \textbf{3.14} & 5.12 & 26.61 & 3.52 & 0.620 \\
20 & 4.42 & 25.81 & 3.01 & 4.76 & 27.12 & 3.94 & \textbf{4.87} & \textbf{26.34} & \textbf{3.19} & 0.838 \\
\bottomrule
\end{tabular}
}
\label{tab:2D_pose_test}
\end{table*}

\subsubsection{Quantitative Evaluation of NeurIK} 
As per Table \ref{tab:sota_results}, the proposed MQ-HMR + NeurIK method introduces significant advancements in biomechanically accurate 3D human pose estimation, as demonstrated across the BML-MoVi, BEDLAM, and OpenCap datasets. By focusing on key metrics like joint angle and marker position accuracy—crucial for biomechanical applications—our model outperforms previous approaches.  On the BML-MoVi dataset, the joint angle error (MAE$_{\textit{angle}}$) is reduced by 14.2\%, and marker position error (MPBLPE) improves by 28.3\%, underscoring the model’s precision in estimating detailed biomechanical movements. The improvement is even more pronounced on the BEDLAM dataset, where joint angle error sees an 18.4\% reduction, a testament to the model’s robustness in handling complex synthetic motions. Marker position accuracy also improves by 27.5\%, highlighting the precision of our model in predicting anatomically correct body movements even in challenging environments.  For OpenCap dataset, our method still delivers a 6.4\% reduction in joint angle error, compared with HMR2.0 + NeurIk baseline, and an impressive 29\% reduction compared with the OpenCap multi-camera tracking system, a clinically-proved motion analysis solution for biomechanical and healthcare applications. 

\vspace{-4pt}

\section{Ablation Study}



\subsection{Effectiveness of Pose Query Token}The Effectiveness of Pose Query Tokens in MQ-HMR highlights the significant influence of token quantity on model performance across datasets as shown in Table \ref{tab:pose_tokens}. Starting with 4 tokens, the model underperforms, showing an MPJPE of 98.8 mm on EMDB. As the pose token count increases, there is a clear improvement, with 96 tokens yielding the best results: an MPJPE of 69.0 mm and MVE of 79.8 mm on 3DPW, and 98.9 mm on EMDB. However, beyond 96 tokens, the improvements diminish. At 192 tokens, MPJPE increases to 74.7 mm, and performance plateaus or declines further with 384 tokens. This indicates that 96 tokens strike the optimal balance between maximizing accuracy and maintaining computational efficiency.

\begin{table}[h!]
\caption{Impact of Pose Query Token on MQ-HMR}
\vspace{-7pt}
\centering
\scalebox{0.7}{
\begin{tabular}{cccccc}
\toprule
 & \multicolumn{1}{c}{H36M} & \multicolumn{2}{c}{3DPW} & \multicolumn{2}{c}{EMDB} \\ \hline
\makecell{\# of \\
Pose Tokens} & \makecell{MPJPE} & \makecell{MPJPE} & \makecell{MVE} & \makecell{MPJPE} & \makecell{MVE } \\ 
\toprule
4 & 44.4 & 75.3 & 85.7 & 98.8 & 103.7 \\
24 & 43.9 & 73.6 & 83.9 & 95.5 & 102.3 \\
48 & 45.1 & 73.1 & 83.2 & 95.5 & 100.4 \\
96 & \textbf{42.5} & \textbf{69.0} & \textbf{79.8} & \textbf{92.5} & \textbf{98.9} \\
192 & 44.5 & 74.7 & 84.6 & 95.9 & 101.6 \\
384 & 44.1 & 74.1 & 84.5 & 95.7 & 99.7 \\
\bottomrule
\end{tabular}
}
\label{tab:pose_tokens}
\end{table}

\subsection{Impact of Feature Resolutions} The ablation study of MQ-HMR reveals clear trends in how multi-scale feature resolutions impact 3D human pose estimation performance as shown in Table \ref{tab:feat_scale}. As feature resolutions increase from \(1\times\) to \(16\times\), the MPJPE consistently improves, reducing errors by 3.5 mm on 3DPW and 2.9 mm on EMDB. Including intermediate scales such as \(4\times\) and \(8\times\) further reduces errors, showing that the model benefits from a balance of both global and local information. Conversely, when lower resolutions (\(1\times\) or \(4\times\)) are excluded, errors increase sharply, particularly on 3DPW, where MPJPE rises by 6.1 mm, indicating the importance of low-resolution features for maintaining overall pose structure. These trends emphasize the effectiveness of multi-scale feature fusion in MQ-HMR, balancing fine detail and broader spatial context for optimal performance.

\begin{table}[h!]
\caption{Impact of feature resolutions on MPJPE error on 3DPW and EMDB datasets. \textcolor{green}{\ding{51}} indicates inclusion, \textcolor{red}{\ding{55}} indicates exclusion, with deltas ($\Delta$) showing error differences with all feature maps.}
\vspace{-7pt}
\centering
\scalebox{0.8}{
\begin{tabular}{cccccc}
\toprule
\multicolumn{4}{c}{Feature Scale} & \multicolumn{2}{c}{MPJPE (↓)} \\
\cmidrule(lr){1-4} \cmidrule(lr){5-6}
1$\times$ & 4$\times$ & 8$\times$ & 16$\times$ & 3DPW & EMDB \\
\midrule
\textcolor{green}{\ding{51}} & \textcolor{green}{\ding{51}} & \textcolor{green}{\ding{51}} & \textcolor{green}{\ding{51}} & 69.0 & 91.5 \\
\textcolor{green}{\ding{51}} & \textcolor{red}{\ding{55}} & \textcolor{red}{\ding{55}} & \textcolor{red}{\ding{55}} & 75.1$_{\Delta 6.1}$ & 95.5$_{\Delta 4.0}$ \\
\textcolor{green}{\ding{51}} & \textcolor{green}{\ding{51}} & \textcolor{red}{\ding{55}} & \textcolor{red}{\ding{55}} & 73.5$_{\Delta 4.5}$ & 93.4$_{\Delta 1.9}$ \\
\textcolor{green}{\ding{51}} & \textcolor{green}{\ding{51}} & \textcolor{green}{\ding{51}} & \textcolor{red}{\ding{55}} & 72.5$_{\Delta 3.5}$ & 92.9$_{\Delta 1.4}$ \\
\textcolor{green}{\ding{51}} & \textcolor{green}{\ding{51}} & \textcolor{green}{\ding{51}} & \textcolor{green}{\ding{51}} & 69.0$_{\Delta 0.0}$ & 92.5$_{\Delta 1.0}$ \\
\textcolor{red}{\ding{55}} & \textcolor{green}{\ding{51}} & \textcolor{green}{\ding{51}} & \textcolor{green}{\ding{51}} & 70.6$_{\Delta 1.6}$ & 90.8$_{\Delta 0.7}$ \\
\textcolor{green}{\ding{51}} & \textcolor{red}{\ding{55}} & \textcolor{green}{\ding{51}} & \textcolor{green}{\ding{51}} & 71.5$_{\Delta 2.5}$ & 91.7$_{\Delta 1.2}$ \\
\textcolor{green}{\ding{51}} & \textcolor{green}{\ding{51}} & \textcolor{red}{\ding{55}} & \textcolor{green}{\ding{51}} & 72.3$_{\Delta 3.3}$ & 91.4$_{\Delta 0.9}$ \\
\textcolor{green}{\ding{51}} & \textcolor{green}{\ding{51}} & \textcolor{green}{\ding{51}} & \textcolor{red}{\ding{55}} & 70.1$_{\Delta 1.1}$ & 90.5$_{\Delta 1.0}$ \\
\bottomrule
\end{tabular}
}
\label{tab:feat_scale}
\end{table}

\subsection{Effectiveness of 2D Pose-Informed Refinement}


Table \ref{tab:2D_pose_test} shows the effect of varying iteration numbers on the 2D pose-informed model across the BML-MoVi, BEDLAM, and OpenCap datasets. Performance generally improves with more iterations, peaking at 10 iterations for BML-MoVi and BEDLAM, where the model achieves the lowest (MAE${_\textit{body}}$ and MAE${_\textit{angle}}$, 3.97 and 2.84 for BML-MoVi, and 4.28 and 3.14 for BEDLAM). However, beyond 10 iterations, performance slightly declines, showing diminishing returns. For the OpenCap dataset, performance remains optimal at 20 iterations, with MAE${_\textit{body}}$ of 4.87 and MAE${_\textit{angle}}$ of 3.19. This underscores the method's ability to achieve significant improvements in just a few iterations while balancing computational time. For example, at 5 iterations, the model performs well (MAE${_\textit{body}}$ of 4.12 on BML-MoVi) with an AITI of just 0.405 seconds (NVIDIA Quadro RTX A4000). This shows that effective refinement is achieved early, making the approach highly efficient for real-time applications.

\vspace{-4pt}

\section{Conclusion} 
We introduce BioPose, a framework for biomechanically accurate 3D human pose estimation from monocular video. BioPose integrates human mesh recovery with biomechanical analysis, combining Multi-Query Human Mesh Recovery (MQ-HMR) for mesh reconstruction, Neural Inverse Kinematics (NeurIK) for biomechanical refinement, and 2D-informed pose alignment. Experiments show its significant gains over state-of-the-art methods, making it a powerful tool for clinical assessments, sports analysis, and rehabilitation technology.
{\small
\bibliographystyle{ieee_fullname}
\bibliography{egbib}
}
\input{supplementary}

\end{document}

%% file: supplementary.tex
\clearpage

\setcounter{table}{0}  
\setcounter{figure}{0} 
\setcounter{section}{0}
\section{Appendix}

\subsection{Overview}

The appendix is organized into the following sections:

\begin{itemize}
    \item Section \ref{sec:Implementation}: Implementation Details
    \item Section \ref{sec:Datasets}: Datasets
    \item Section \ref{sec:Evaluation Metrics}: Evaluation Metrics
    \item Section \ref{sec:Augmentation}: Data Augmentation
    \item Section \ref{sec:Camera}: Camera Model 
    \item Section \ref{sec:Backbones}: Impact of Backbones
    \item Section \ref{sec:Deformable}: Impact of Deformable Cross-Attention Layers
    \item Section \ref{sec:Losses_effects}: Effect of losses on NeurIK model
    \item Section \ref{sec:MultiFrame}: Multi-frame out vs Single frame out
    \item Section \ref{sec:FramesNeurIK}: Impact of Number of Frames in NeuralIK
    \item Section \ref{sec:Qualitative}: Qualitative Results

\end{itemize}

\subsection{Implementation Details}
\label{sec:Implementation}
\partitle{MQ-HMR} Our MQ-HMR model is implemented in PyTorch. To achieve this, we utilized multi-resolution feature maps at 4×, 8×, and 16× scales, ensuring the model captures both local detail and global structure. The total loss function in MQ-HMR is defined as $\mathcal{L}_{\text{total}} = \mathcal{L}_{\text{SMPL}} + \mathcal{L}_{\text{3D}} + \mathcal{L}_{\text{2D}}$. This combines 3D loss ($\mathcal{L}_{\text{3D}}$), 2D loss ($\mathcal{L}_{\text{2D}}$), and SMPL parameter loss ($\mathcal{L}_{\text{SMPL}}$) to optimize the shape ($\beta$) and pose ($\theta$) parameters in the SMPL space. The loss weights for this stage are carefully tuned to balance each objective, where the pose component is weighted at $\lambda_{\theta}=1 \times 10^{-3}$ and the shape component at $\lambda_{\beta}=5 \times 10^{-4}$. For the 3D loss, $\lambda_{\text{3D}}=5 \times 10^{-2}$, and for the 2D loss, $\lambda_{\text{2D}}=1 \times 10^{-2}$. The architecture incorporates 96 Pose Token Queries ($Q$) and 4 deformable cross-attention layers as default, which enables the model to attend to relevant spatial information across different scales. The model was trained for 100K iterations using the Adam optimizer with a batch size of 48 and a learning rate of $1 \times 10^{-5}$.

\partitle{NeurIK} Our NeurIK module is implemented in PyTorch and processes 
virtual markers  $\mathbf{X}_{\text{VM}}^{\text{exp}} \in \mathbb{R}^{142 \times 3}$  extracted from the 3D mesh produced by MQ-HMR. The architecture includes a Spatial Convolution Encoder that utilizes 1D convolutional layers for spatial feature extraction and a Temporal Transformer Encoder that employs multi-head self-attention to model temporal dependencies across multiple frames. The total loss function is defined as \( \mathcal{L}_{\text{neurIK}} = \lambda_j \mathcal{L}_{\text{j}} + \lambda_m \mathcal{L}_{\text{m}} + \lambda_s \mathcal{L}_{\text{s}} + \lambda_q \mathcal{L}_{\text{q}} \), with the weights set as \( \lambda_j = 1.0 \), \( \lambda_m = 2.0 \), \( \lambda_s = 0.1 \), and \( \lambda_q = 0.06 \). The model was trained for 25 epochs using the Adam optimizer with an initial learning rate of 0.001, decaying to \( 5 \times 10^{-6} \), and a batch size of 128. Data augmentation techniques such as scaling, rotation, translation, and noise injection were applied to increase the model’s robustness to occlusions and real-world variations.

To generate the input for our network, the SMPL mesh for each video frame was recovered using a test-time optimized HMR 2.0 model. The vertex indices of virtual markers on the SMPL mesh were used to calculate marker locations, which served as critical inputs for the subsequent spatio-temporal modeling. During training, we employed data augmentation methods such as scaling, rotation, translation, and noise to mimic occlusions, following the approach used in \cite{bittner2022towards}. 
Our model was trained using the following hyperparameters and loss functions. We used an Adam optimizer with a weight decay of 0.001 and a batch size of 128. The learning rate decayed exponentially from an initial rate of 0.001 to a final rate of $5 \times 10^{-6}$ over 25 epochs. For the spatio-temporal model, we set the hyperparameters experimentally, adjusting key parameters as needed throughout the training process.

\subsection{Datasets} 
\label{sec:Datasets}

We utilized videos from BMLmovi \cite{ghorbani2021movi}, OpenCap\cite{uhlrich2023opencap} and BEDLAM \cite{black2023bedlam}datasets.
We trained our NeurIK model on BMLmovi data and tested on OpenCap and BEDLAM datasets. 


\textbf{BML-MoVi:}  BMLMovi consists of 90 subjects performing 21 different actions, captured using two cameras and a marker-based motion capture system. As BMLMovi lacks kinematic annotations, we processed the available marker data through the OpenSim Scale and OpenSim IK tools to accurately determine joint angles and body segment dimensions for ground truth measurements. \cite{ghorbani2021movi}.

\textbf{OpenCap:} OpenCap includes data from ten subjects performing various actions such as walking, squatting, standing up from a chair, drop jumps, and their asymmetric variations. The recordings were made using five RGB cameras alongside a marker-based motion capture system. Additionally, OpenCap offers processed marker data and kinematic annotations for a comprehensive full-body OpenSim skeletal model. 

\textbf{BEDLAM:} BEDLAM dataset comprises synthetic video data featuring a total of 271 subjects, including 109 men and 162 women. It includes monocular RGB videos paired with ground-truth 3D bodies in SMPL-X format, covering a diverse range of body shapes, skin tones, and motions. The dataset features realistic animations with detailed hair and clothing simulated using physics, enhancing the realism of the data.  To obtain kinematic annotations, we leverage the vertices of SMPL-X mesh to generate virtual markers and processed the available marker data through the OpenSim Scale and OpenSim IK tools to  determine joint angles and body segment dimensions as the ground-truth data. \cite{black2023bedlam}.

\subsection{Evaluation Metrics}
\label{sec:Evaluation Metrics}

For NeurIK, we employ metrics that focus on biomechanical accuracy. These include Mean Per Bony Landmarks Position Error (MPBLPE), which measures the accuracy of predicted bony landmarks against ground truth positions, and Mean Absolute Error for body scale (MAE$_{\textit{body}}$), which evaluates the correctness of predicted body segment dimensions by comparing their longest axes in millimeters. Additionally, we report Mean Absolute Error for joint angles (MAE$_{\textit{angle}}$), which assesses the precision of joint angle predictions in degrees, critical for biomechanical simulations such as joint force and muscle force analysis.

\textbf{MPBLPE}\textbf{:} Mean Per Bony Landmarks Position Error, measures the accuracy of predicted bony landmark positions by comparing them to ground truth data. This metric is inspired by the mean per joint position error (MPJPE) which is often used in 3D pose estimation. MPBLPE involves aligning both the predicted and actual positions at a common root point and then calculating the average Euclidean distance between corresponding landmarks, i.e.,
\[
\text{MPBLPE}(\text{pred}, \text{target}) = \frac{1}{N} \sum_{i=1}^{N} \sqrt{\sum_{j=1}^{3} (\text{pred}_{ij} - \text{target}_{ij})^2}
\]
where \( i \) indexes the joints, and \( j \) indexes the spatial dimensions (X, Y, Z). For each joint \( i \), \( \text{pred}_{ij} \) and \( \text{target}_{ij} \) represent the predicted and target positions in dimension \( j \), respectively. The entire expression is divided by \(N\), the number of bony landmarks, to calculate the average Euclidean distance per landmark, which normalizes the loss to account for differences in the number of joints among different datasets or models.

\textbf{MAE}$_{\textit{ body}}$\textbf{:} The axis corresponding to the longest dimension of each body segment is selected, and its scale is converted into millimeters to calculate the Mean Absolute Error (MAE$_{\textit{ body}}$). Specifically, the x-axis is used for the skull, toes, and calcaneus; the y-axis is chosen for the spine, lower limbs, and upper limbs; the z-axis is applied for the jaw, scapula, and clavicle; and for the pelvis, all three axes are considered \cite{bittner2022towards}.

\textbf{MAE}$_{\textit{ angle}}$\textbf{:} MAE$_{\textit{ angle}}$ represents the mean absolute error of the joint angle, measured in degrees \cite{bittner2022towards,uhlrich2023opencap}.

In this study, MAE$_{\textit{ angle}}$ is prioritized over MPBLPE because joint angles, rather than marker positions, will be utilized in subsequent applications of musculoskeletal models, such as simulating joint reaction forces and analyzing muscle forces. Consequently, ensuring the accuracy of joint angles is crucial for producing precise force simulations, making it more significant than the accuracy of marker positions.

\subsection{Data Augmentation}
\label{sec:Augmentation}

To enhance MQ-HMR's robustness and generalization, we applied extensive data augmentation during training. This included random scaling, rotation, horizontal flips, and color jittering on both images and poses. These augmentations help the model better handle real-world challenges like occlusions and incomplete body information. As a result, the data augmentation process significantly contributes to MQ-HMR's improved performance in human mesh reconstruction by making the model more adaptable and resilient to diverse and unpredictable inputs.

\subsection{Camera Model} 
\label{sec:Camera}

In our approach, MQ-HMR utilizes a weak perspective camera model with a fixed focal length 5000 and an intrinsic matrix $K \in \mathbb{R}^{3 \times 3}$. To simplify the computation, the rotation matrix $R$ is set to the identity matrix $I_3$, allowing us to focus solely on the translation vector $T \in \mathbb{R}^3$. The 3D joints $J_{\text{3D}}$ are then projected onto 2D coordinates $J_{\text{2D}}$ using the equation $J_{\text{2D}} = \Pi(K(J_{\text{3D}} + T))$, where $\Pi$ denotes the perspective projection based on camera intrinsics $K$. This simplification reduces the number of parameters involved, improving the computational efficiency of our human mesh recovery process.

\subsection{Impact of Backbones}
\label{sec:Backbones}

The comparison of backbones in MQ-HMR highlights that the ViT-H backbone consistently delivers superior performance in terms of MPJPE on both the 3DPW and EMDB datasets (Table \ref{tab:backbones}). Specifically, ViT-H achieves the lowest MPJPE, with 69.0 mm on 3DPW and 92.5 mm on EMDB, indicating its strength in accurate pose estimation. While HRNet-w48 shows a slight advantage in MVE on 3DPW, with 79.5 mm compared to ViT-H's 79.8 mm, this minor improvement in mesh vertex error does not offset its higher MPJPE. On the more complex EMDB dataset, ViT-H maintains a clear edge in both MPJPE and MVE, further solidifying its effectiveness in capturing detailed pose structures. Thus, ViT-H is the more robust backbone for 3D human pose estimation, especially when precision across both metrics is essential.

\begin{table}[h!]
\caption{Impact of Backbones on MQ-HMR}
\vspace{-7pt}
\centering
\scalebox{0.85}{
\begin{tabular}{ccccc}
\toprule
 & \multicolumn{2}{c}{3DPW} & \multicolumn{2}{c}{EMDB} \\ \hline
\makecell{} & \makecell{MPJPE \\ (↓)} & \makecell{MVE \\ (↓)} & \makecell{MPJPE \\ (↓)} & \makecell{MVE \\ (↓)} \\ 
\toprule
ViT-H  96 & \textbf{69.0} & 79.8 & \textbf{92.5} & \textbf{98.9} \\
HRNet-w48 & 70.3 & \textbf{79.5} & 93.5 & 100.5 \\
\bottomrule
\end{tabular}
}
\label{tab:backbones}
\end{table}

\subsection{Impact of Deformable Cross-Attention Layers}
\label{sec:Deformable}

The impact of deformable cross-attention layers in MQ-HMR reveals that the optimal number of layers for accurate 3D pose estimation is 4 (Table \ref{tab:cross_attention_layers}). With 4 layers, the model achieves the lowest MPJPE on both the 3DPW (69.0 mm) and EMDB (92.5 mm) datasets, capturing pose information effectively across different complexities. Increasing the number of layers to 6 offers a slight improvement in mesh vertex error (MVE), particularly reducing it to 78.9 mm on 3DPW, but at the cost of a higher MPJPE. However, further increasing the number of layers to 8 results in diminished performance, as seen with the 72.4 mm MPJPE for 3DPW, suggesting that too many layers may introduce redundancy and inefficiencies. Additionally, the higher number of cross-attention layers increases the computational burden without significant accuracy gains. Therefore, 4 deformable cross attention layers provide the best balance between computational efficiency and performance, avoiding the computationally heavy overhead of higher-layer configurations.

\begin{table}[h!]
\caption{Impact of \# of Deformable Cross Attention Layers in MQ-HMR. }
\vspace{-7pt}
\centering
\scalebox{0.85}{
\begin{tabular}{ccccc}
\toprule
 & \multicolumn{2}{c}{3DPW} & \multicolumn{2}{c}{EMDB} \\ \hline
\# of Deformable \\  Cross Attention \\  Layers & \makecell{MPJPE} & \makecell{MVE} & \makecell{MPJPE} & \makecell{MVE} \\ 
\toprule
2 & 75.2 & 87.1 & 98.1 & 105.6 \\
4 & \textbf{69.0} & 79.8 & \textbf{92.5} & \textbf{98.9} \\
6 & 70.4 & \textbf{78.9} & 93.1 & 103.1 \\
8 & 72.4 & 84.1 & 93.5 & 102.5 \\
\bottomrule
\end{tabular}
}
\label{tab:cross_attention_layers}
\end{table}

\subsection{Effect of Losses on NeurIK Model}
\label{sec:Losses_effects}

We assess the effect of various loss terms as presented in Table \ref{tab:loss_results}, by progressively introducing each loss during the model training process. Initially, the model is trained using only \(L_{\text{m}}\) (marker loss), followed by the sequential addition of \(L_{\text{j}}\) (joint loss), \(L_{\text{q}}\) (angle loss), and finally \(L_{\text{s}}\) (body scale loss). The results show that training with only \(L_{\text{m}}\) achieves the lowest MPBLPE (21.57), indicating good alignment of bony landmarks. However, this configuration yields relatively high errors in body scale (MAE\(_{\text{body}}\) = 8.46) and joint angles (MAE\(_{\text{angle}}\) = 7.58), highlighting its limitations in capturing accurate body dimensions and angles.  Introducing \(L_{\text{j}}\) into the training improves both the body and angle predictions, reducing MAE\(_{\text{body}}\) to 7.18 and MAE\(_{\text{angle}}\) to 6.43, although MPBLPE slightly increases to 22.26. Further, the inclusion of \(L_{\text{q}}\) significantly improves joint angle accuracy, achieving the lowest MAE\(_{\text{angle}}\) (2.34), though this comes at the cost of a slight increase in body scale error (MAE\(_{\text{body}}\) = 6.21). Finally, adding \(L_{\text{s}}\) results in the best body scale accuracy, with MAE\(_{\text{body}}\) reduced to 3.97, though the joint angle error slightly increases to MAE\(_{\text{angle}}\) = 2.84. These findings demonstrate that while each loss term optimizes different aspects of the model's performance, utilizing all four losses together achieves a balanced improvement across both body scales and joint angles.

\begin{table}[htbp]
\centering
\caption{Effect of losses on NeurIK model}
\begin{tabular}{lcccc}
\toprule
 & \(L_{\text{m}}\) & \(L_{\text{j}}\) & \(L_{\text{q}}\) & \(L_{\text{s}}\) \\
\midrule
\(L_{\text{m}}\) & \(\checkmark\) & \(\checkmark\) & \(\checkmark\) & \(\checkmark\) \\
\(L_{\text{j}}\)  &  & \(\checkmark\) & \(\checkmark\) & \(\checkmark\) \\
\(L_{\text{q}}\)  &  &  & \(\checkmark\) & \(\checkmark\) \\
\(L_{\text{s}}\)   &  &  &  & \(\checkmark\) \\
\midrule
MPBLPE & \textbf{21.57} & 22.26 & 24.08 & 25.76 \\
MAE$_{\text{body}}$ & 8.46 & 7.18 & 6.21 & \textbf{3.97} \\
MAE$_{\text{angle}}$ & 7.58 & 6.43 & \textbf{2.34} & 2.84 \\
\bottomrule
\end{tabular}

\label{tab:loss_results}
\end{table}

\subsection{Multi-frame out vs Single frame out}
\label{sec:MultiFrame}

Table 4 compares the performance of NeurIK using two different temporal models: multiple frame out and single last frame out, across three datasets: BML-MoVi, BEDLAM, and OpenCap. In the multi-frame temporal model, instead of predicting just the last frame in a sequence, the model predicts all 64 frames. The results show that the single last frame out model consistently outperforms the multiple frame out model across all metrics and datasets. For instance, in BML-MoVi, the single frame out model achieves lower MAE${_\textit{body}}$ (3.97 vs. 4.01) and MAE${_\textit{angle}}$ (2.84 vs. 2.95). Similar improvements are seen in BEDLAM and OpenCap, suggesting that the single frame out model offers better accuracy, making it the preferred choice for NeurIK.
The single frame out model likely outperforms the multiple frame out model for several reasons. First, by focusing solely on predicting the final frame, the model can concentrate its capacity on optimizing that specific output, resulting in more precise predictions. In contrast, the multiple frame out model must predict the entire sequence, which can introduce cumulative error across frames. Additionally, predicting all frames may create temporal inconsistencies, as the model tries to maintain coherence throughout the sequence. The single frame out model avoids these challenges, offering a simplified learning objective that reduces complexity and leads to better performance, especially in terms of MAE. This makes it a more efficient choice for tasks like NeurIK, where accuracy in the final frame is critical.
\begin{table*}[h!]
\centering
\caption{Impact of different temporal model on NeurIK}
\scalebox{0.75}{
\begin{tabular}{l|ccc|ccc|ccc}
\toprule
 & \multicolumn{3}{c|}{BML-MoVi} & \multicolumn{3}{c|}{BEDLAM} & \multicolumn{3}{c}{OpenCap} \\
Temporal model & \makecell{MAE$_{\textit{body}}$} & \makecell{MPBLPE} & \makecell{MAE$_{\textit{angle}}$} & \makecell{MAE$_{\textit{body}}$} & \makecell{MPBLPE} & \makecell{MAE$_{\textit{angle}}$} & \makecell{MAE$_{\textit{body}}$} & \makecell{MPBLPE} & \makecell{MAE$_{\textit{angle}}$} \\
\midrule
Multiple frame out & 4.01 & 25.84 & 2.95 & 4.41 & 26.61 & 3.32 & 5.12 & 26.72 & 3.44 \\

Single frame out & \textbf{3.97} & \textbf{25.76} & \textbf{2.84} & \textbf{4.28} & \textbf{26.54} & \textbf{3.14} & \textbf{4.87} & \textbf{26.34} & \textbf{3.19} \\

\bottomrule
\end{tabular}
}
\end{table*}

\begin{table*}[h!]
\centering
\caption{Impact of number of frames in temporal model on NeurIK}
\scalebox{0.75}{
\begin{tabular}{l|ccc|ccc|ccc}
\toprule
 & \multicolumn{3}{c|}{BML-MoVi} & \multicolumn{3}{c|}{BEDLAM} & \multicolumn{3}{c}{OpenCap} \\
\#\ of Frames & \makecell{MAE$_{\textit{body}}$} & \makecell{MPBLPE} & \makecell{MAE$_{\textit{angle}}$} & \makecell{MAE$_{\textit{body}}$} & \makecell{MPBLPE} & \makecell{MAE$_{\textit{angle}}$} & \makecell{MAE$_{\textit{body}}$} & \makecell{MPBLPE} & \makecell{MAE$_{\textit{angle}}$} \\
\midrule
16 & 5.08 & 26.83 & 4.12 & 5.76 & 27.62 & 4.35 & 5.98 & 27.82 & 4.56 \\
32 & 4.52 & 26.28 & 3.62 & 5.23 & 27.19 & 3.82 & 5.53 & 27.15 & 3.97 \\
64 & \textbf{3.97} & \textbf{25.76} & \textbf{2.84} & \textbf{4.28} & \textbf{26.54} & \textbf{3.14} & \textbf{4.87} & \textbf{26.34} & \textbf{3.19} \\
128 & 4.67 & 26.41 & 3.43 & 5.15 & 27.12 & 3.94 & 5.52 & 27.04 & 4.13 \\
\bottomrule
\end{tabular}
}
\label{tab:neurik_frames}

\end{table*}

\subsection{Impact of Number of Frames in NeuralIK}
\label{sec:FramesNeurIK}

Table 5 illustrates the impact of varying the number of frames used in the temporal model on the performance of NeurIK across three datasets: BML-MoVi, BEDLAM, and OpenCap. The results show that the number of frames significantly affects the accuracy of the model, with optimal performance generally observed at 64 frames across all datasets. For the BML-MoVi dataset, increasing the number of frames from 16 to 64 leads to a consistent reduction in both MAE${_\textit{body}}$ and MAE${_\textit{angle}}$, with the lowest errors achieved at 64 frames (3.97 for MAE${_\textit{body}}$ and 2.84 for MAE${_\textit{angle}}$). Similarly, the MPBLPE decreases as the number of frames increases, reaching 25.76 at 64 frames. However, performance begins to degrade at 128 frames, where both MAE${_\textit{body}}$ and MAE${_\textit{angle}}$ increase. A similar trend is observed in the BEDLAM and OpenCap datasets. For BEDLAM, the MAE${_\textit{body}}$ reduces from 5.76 at 16 frames to 4.28 at 64 frames, with a corresponding improvement in MAE${_\textit{angle}}$ (from 4.35 to 3.14). In the OpenCap dataset, the optimal performance is also achieved at 64 frames, with MAE${_\textit{body}}$ reaching 4.87 and MAE${_\textit{angle}}$ improving to 3.19. This ablation study shows that 64 frames strike the right balance between capturing enough temporal information and maintaining computational efficiency. While performance improves when increasing frames from 16 to 64, using 128 frames offers no further gains and even slightly degrades performance. The likely reason is that 64 frames provide sufficient motion dynamics and biomechanical patterns for accurate pose estimation, while fewer frames lack context, and more frames introduce redundant information. Thus, 64 frames offer the optimal amount of temporal data without adding unnecessary complexity or noise.

\subsection{Qualitative Results}
\label{sec:Qualitative}

\begin{figure*}[ht] 
    \centering
\includegraphics[width=0.7\linewidth]{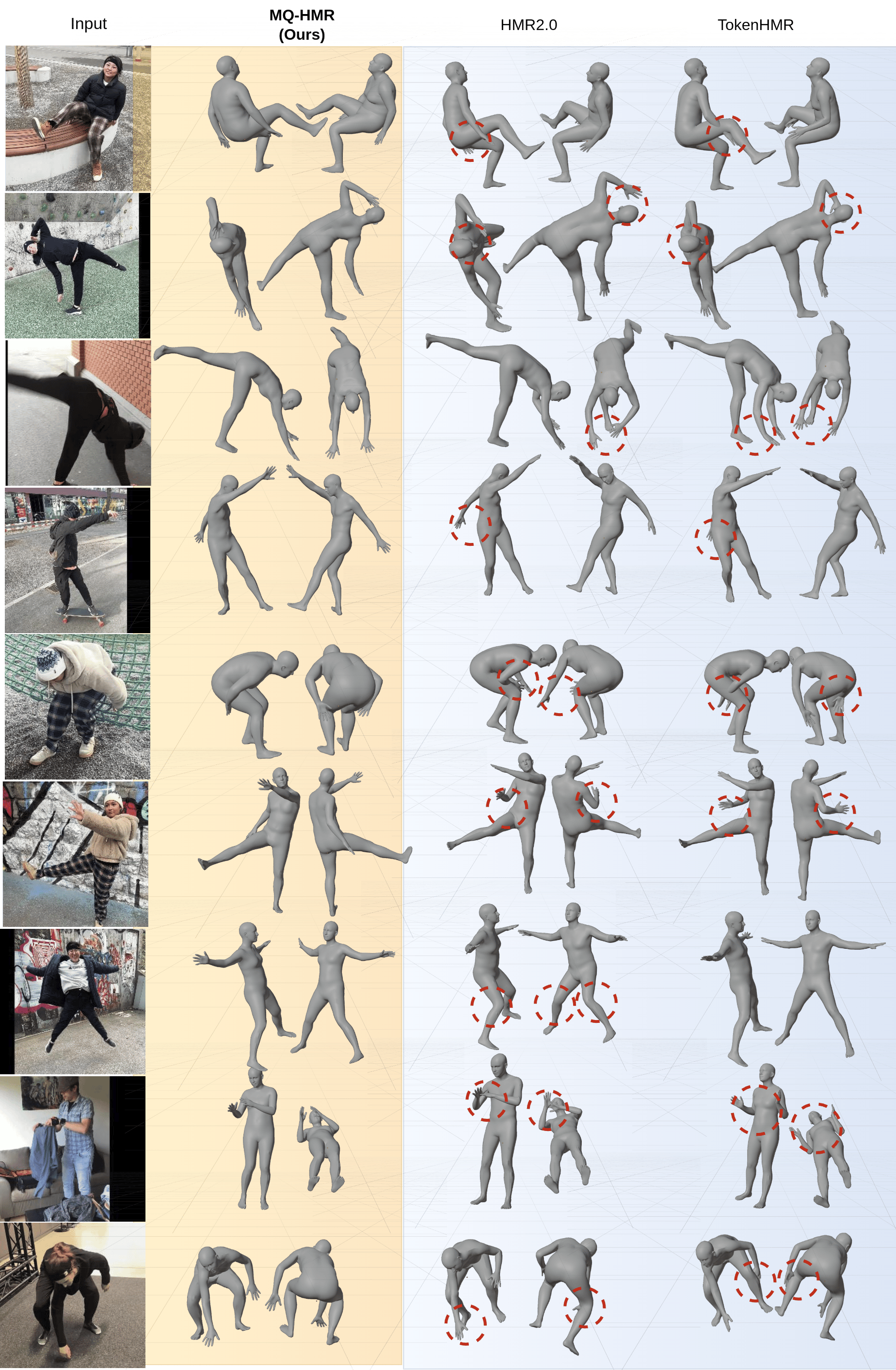}
\caption{Comparison of state-of-the-art methods, HMR2.0 \cite{goel2023humans} and TokenHMR \cite{dwivedi2024tokenhmr}, which use vision transformers for 3D human mesh recovery from a single image. Red circles highlight errors in these methods when dealing with complex or ambiguous poses. In contrast, our MQ-HMR method addresses these challenges by incorporating a multi-query deformable transformer, leveraging multi-scale feature maps and a deformable attention mechanism to deliver more accurate and anatomically consistent pose estimations, even in difficult scenarios.}
\label{fig:mqhmr_sota}
\end{figure*}

\begin{figure*}[ht]
    \centering
    \includegraphics[width=0.85\linewidth]{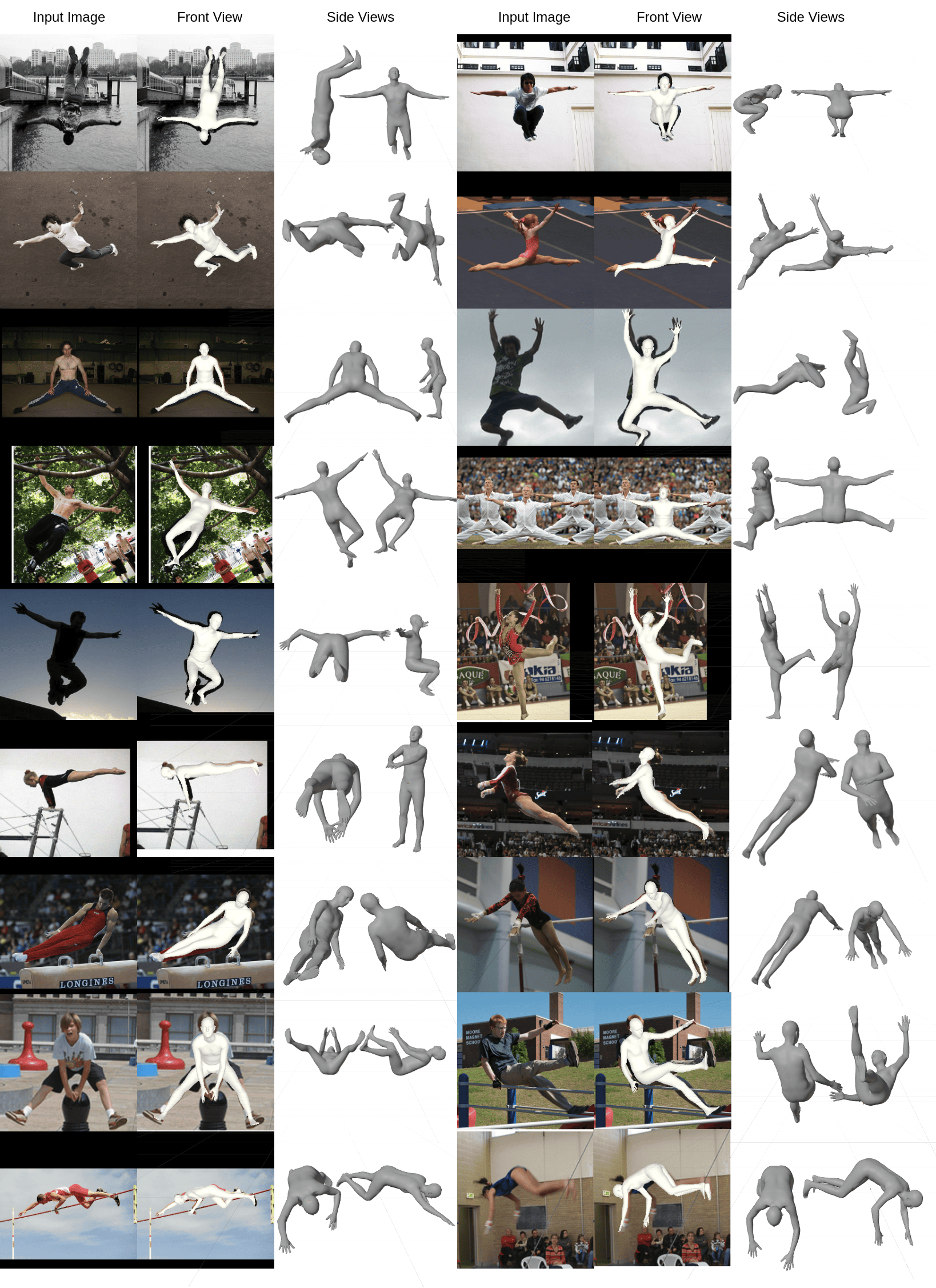}
\caption{Qualitative results of our approach on challenging poses from the LSP \cite{johnson2011learning} dataset.}
\label{fig:mqhmr_chall_poses_3d}
\end{figure*}

\begin{figure*}[ht] 
    \centering
    \includegraphics[width=1.0\linewidth]{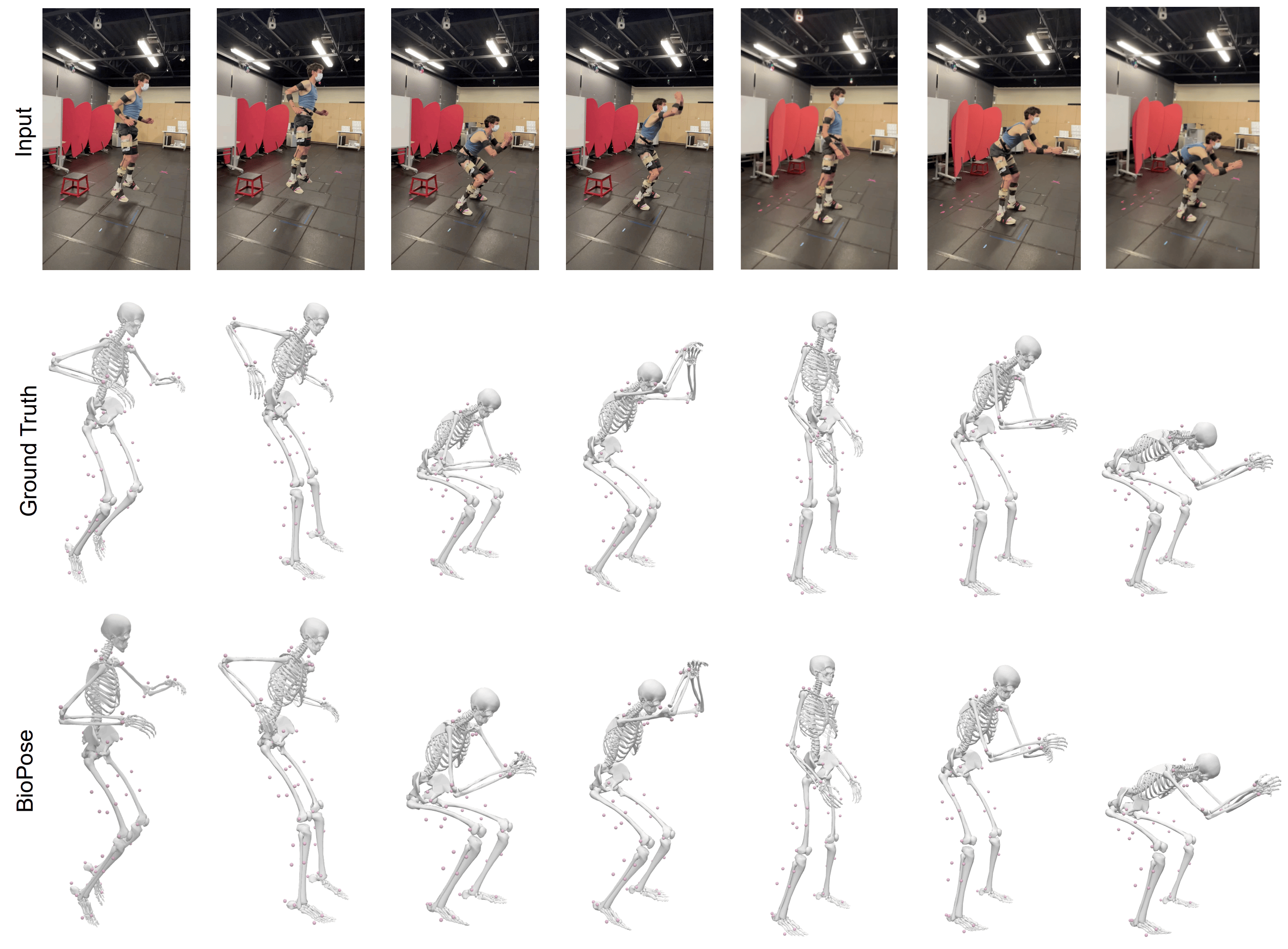}
\caption{Qualitative results of our proposed method BioPose and comparison with ground truth. These pictures include multiple actions such as squatting and drop jumps.}
\label{fig:BioPose-D3KE}
\end{figure*}

We present qualitative results of MQ-HMR in Figures \ref{fig:mqhmr_sota} and \ref{fig:mqhmr_chall_poses_3d}, showcasing the model's capability in handling extreme poses and partial occlusions. These results demonstrate the effectiveness of MQ-HMR, where the 3D reconstructions align well with the input images and maintain accuracy when viewed from different perspectives. A key factor behind this success is MQ-HMR's multi-query deformable attention mechanism, which efficiently manages uncertainty during the 2D-to-3D mapping process. MQ-HMR is able to overcome challenges that typically affect other state-of-the-art models. This approach ensures that MQ-HMR produces accurate and consistent 3D reconstructions, even in complex or ambiguous scenarios where traditional methods often struggle. Also, We show the qualitative results of BioPose in Figure \ref{fig:BioPose-D3KE}, highlighting how our model is able to predict different poses, very close to the ground truth. The figures show multiple actions like squatting and drop jumping.

\begin{figure*}[ht] 
    \centering
    \includegraphics[width=1\linewidth]{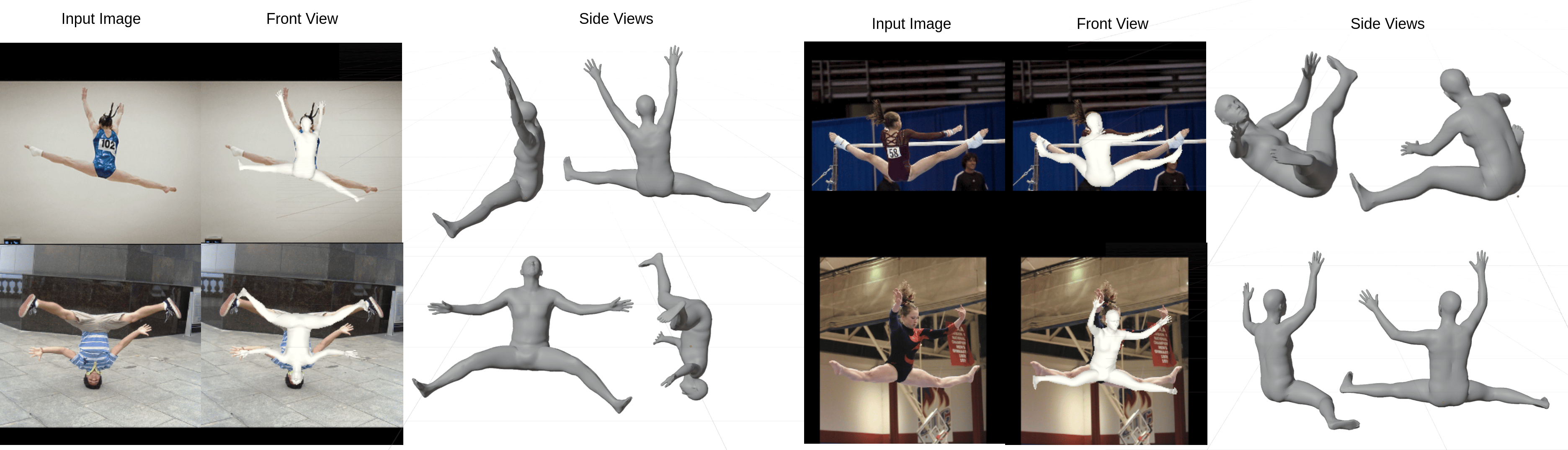}
\caption{Failure Cases of MQ-HMR in 3D Human Reconstruction: MQ-HMR frequently struggles with handling unusual body movements and the complex layering of body parts in three-dimensional space. These difficulties often lead to inaccurate 3D pose estimations and invalid results. The main reason for these limitations is the model's dependence on the SMPL parametric model, which fails to adequately represent the complexity of extreme or rare human pose.}
\label{fig:genhmr_failure}
\end{figure*}